\documentclass[10pt,twocolumn,letterpaper]{article}

\usepackage{cvpr}
\usepackage{times}
\usepackage{epsfig}
\usepackage{graphicx}
\usepackage{amsmath}
\usepackage{amssymb}

\usepackage{subfigure}
\usepackage{amsthm}

\usepackage{multirow}
\usepackage{booktabs}
\usepackage{algorithm}
\usepackage{algorithmic}
\usepackage{epstopdf}
\usepackage{authblk}

\usepackage[pagebackref=true,breaklinks=true,letterpaper=true,colorlinks,bookmarks=false]{hyperref}

 \cvprfinalcopy 

\def\cvprPaperID{812} 

\ifcvprfinal\fi
\begin{document}

\title{Transductive Zero-Shot Hashing via Coarse-to-Fine Similarity Mining}

\author[$^{\dagger}$]{Hanjiang Lai}
\author[$^{\ddagger}$]{Yan Pan \thanks{Corresponding author: Yan Pan, email: panyan5@mail.sysu.edu.cn.}}
\affil[$^{\ddagger}$]{School of Data and Computer Science, Sun Yan-Sen University, China}


\maketitle

\begin{abstract}
Zero-shot Hashing (ZSH) is to learn hashing models for novel/target classes without training data, which is an important and challenging problem. Most existing ZSH approaches exploit transfer learning via an intermediate shared semantic representations between the seen/source classes and novel/target classes. However, due to having disjoint, the hash functions learned from the source dataset are biased when applied directly to the target classes. In this paper, we study the transductive ZSH, i.e., we have unlabeled data for novel classes. We put forward a simple yet efficient joint learning approach via coarse-to-fine similarity mining which transfers knowledges from source data to target data. It mainly consists of two building blocks in the proposed deep architecture: 1) a shared two-streams network, which the first stream operates on the source data and the second stream operates on the unlabeled data, to learn the effective common image representations, and 2) a coarse-to-fine module, which begins with finding the most representative images from target classes and then further detect similarities among these images, to transfer the similarities of the source data to the target data in a greedy fashion. Extensive evaluation results on several benchmark datasets demonstrate that the proposed hashing method achieves significant improvement over the state-of-the-art methods.
\end{abstract}

\section{Introduction}
Due to the ever-growing large-scale image data on the web, image retrieval has attracted increasing interest. Hashing~\cite{LSH} is a popular nearest neighbor search method that learns similarity-preserving hash functions to encode the images into binary codes. 

Many algorithms have been proposed for learning the similarity-preserving hash functions~\cite{ITQ,sh,KSH}. One of the leading approaches is the deep-networks-based hashing~\cite{liu2017deep,zhang2016efficient,li_pairwise}, which learn the powerful image representations as well as the binary hash codes. Lin et al.~\cite{lin2015deep} proposed to learn the hash codes and image representations in a point-wised manner. Li et al.~\cite{li_pairwise} and  Liu et al.~\cite{liu2016deep} presented deep pairwise-supervised hashing. Further, Zhao et al.~\cite{zhao2015deep} presented a deep ranking based method for multi-label images, and Zhuang et al.~\cite{zhuang2016fast} proposed a triplet-based deep hash network.

However, as the image data continue to grow, many new semantic concepts emerge rapidly. And those brand novel classes may emerge with zero or litter labeled images. Thus, it is desirable to learn similarity-preserving hash functions for those novel/target classes. 

Zero-shot learning (ZSL) is to build recognition models that capable of recognising novel classes without labeled training samples. The main idea behind ZSL is to exploit knowledge transfer via embedding both classes into a common semantic representation, e.g., the word-vectors~\cite{mikolov2013distributed} of the class names. Inspired by that, Yang et al.~\cite{yang2016zero} firstly proposed zero-shot hashing (ZSH), which sets up a tunnel to transfer supervised knowledge between the source and target classes via an intermediate-level semantic representation. However, due to source classes and target classes have different data distributions, using the hash functions learned from the source dataset without any adaptation to the target dataset may causes an unknown bias, which is called \textit{projection domain shift problem}~\cite{fu2015transductive,pan2010survey}.  Pachori et al.~\cite{pachori2017hashing} introduced an unsupervised domain-adaptation model for ZSH, which updated the learned hash model by the query data from the target classes. 

\begin{figure*}
\centering
    \includegraphics[width=0.9\hsize \hspace{0.01\hsize}]{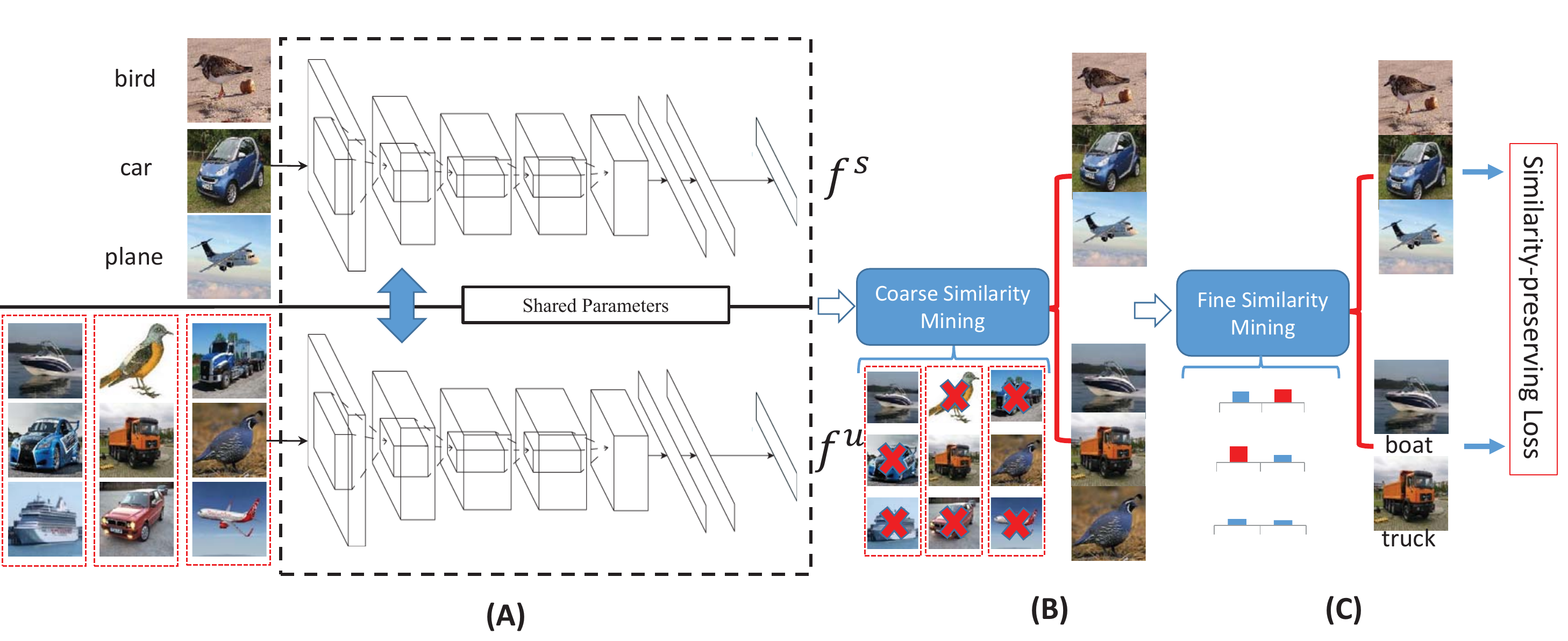}
  \caption{Overview of the proposed deep architecture for the transductive zero-shot hashing. Our architecture is a two-streams network, in which the first stream operates on the labelled images of the seen classes while the second stream operates on the unlabelled images. The architecture firstly (A) encodes the images into the image representations via stacked convolutional and fully-connected layers. With this, it can obtain the common feature representations for both the source and target data. And then it (B) detects the most informative images from the large set of unlabeled images. We propose a greedy binary classification network with the proposed cross-images selection layer to achieve the goal.  (C) is a fine similarity mining module to further finding the similarities among the informative images. It greedily selects the image which has the highest probability score come from a novel class, and then assigns the label of the image as its class.  After that, a similarity-preserving loss function is conducted to learn the hash functions for both the source and target classes. }
  \label{framework}  
\end{figure*}

In this paper, we focus on ZSH with transductive setting~\cite{fu2015transductive,guo2016transductive}, i.e., the unlabelled images from the target classes are available. In the real world, the images of the seen classes are usually more common than the novel ones. It is unrealistic to assume that the seen classes are never existing in the unlabelled dataset. Thus, we do not require the unlabelled data are all from the target classes. Specially, two datasets with disjoint classes are considered in TZSH: a labelled source dataset where all samples are from seen classes and an unlabelled dataset that includes unlabelled images from both of the seen and novel classes.  

We propose deep architecture for transductive zero-shot hashing via coarse-to-fine similarities mining. As shown in Fig.~\ref{framework}, our architecture is a two-streams neural network. The first stream is for labeled images of the seen classes and the second stream for unlabeled images, which are designed to relate the target classes to the source ones. Then, we propose coarse-to-fine similarity mining module to transfer the similarities of the labeled source data to the target data. We begin with the coarse stage which is to find the informative images of the novel classes, i.e., removing the hard samples and noisy images. We formulate it as a binary classification problem, and a new layer called cross-images selection layer is proposed to greedily select the most informative target data. With the images found in the coarse stage, we devise a simple and effective strategy for transferring the similarities from the seen classes to the target classes by utilizing the word representations. Finally, a loss function is proposed to capture the similarities among the images, and a hash model is learned to encode all the images into binary codes.

The main contributions of this work are three-folds.
\begin{itemize}
\item We propose a deep transductive zero-shot hashing framework to solve the projection domain shift problem, which learns the hash functions for both the seen and novel classes. 

\item We propose a simple yet efficient coarse-to-fine similarity mining method for transferring the knowledge from the seen classes to the novel classes.

\item We conduct extensive evaluations on several benchmark datasets. The empirical results demonstrate that the proposed method achieves superior performance to the baselines. 

\end{itemize}

\section{Related Work}
\label{related_work}

Due to the rapidly increasing data, hashing has become a popular method for nearest neighbor search. Many methods have been proposed for hashing, including data independent hashing ~\cite{LSH} and data dependent hashing~\cite{ITQ,AGH,KLSH,sh,semantic,SSH,KSH,MLH,BRE}.

Among the supervised methods, learning the hash codes with deep frameworks, e.g., CNN based methods~\cite{guo2015cnn}, has been emerged as one of the leading approaches. Lin et al.~\cite{lin2015deep} proposed to learn the hash codes and image representations in a point-wised manner, which is suitable for large-scale training datasets. Zhang et al.~\cite{zhang2015bit} presented a novel regularization term to learn the deep hash functions. Wang et al.~\cite{wangdeep} proposed deep multimodal hashing with orthogonal regularization (DMHOR) method for multi-modal data. Zhao et al.~\cite{zhao2015deep} proposed a deep semantic ranking based method for learning hash functions that preserve multi-level semantic similarity between multi-label images. Zhuang~\cite{zhuang2016fast} proposed a fast deep network for triplet supervised hashing. Liu et al.~\cite{liu2016deep} proposed deep supervised hashing (DSH) to learn compact similarity-preserving binary code for a huge body of image data. Zhang et al.~\cite{zhang2016efficient} proposed an efficient training algorithm for very deep neural network by alternating direction method of multipliers. Liu et al.~\cite{liu2017deep} proposed deep sketch hashing (DSH) for free-hand sketch-based image retrieval. Mandal et al.~\cite{nlabel} and Jiang et al.~\cite{jiang2016deep} present deep hashing framework for cross-modal retrieval.

Although the success of the supervised deep hashing methods, they need a lot of label information. Recently, Lin et al.~\cite{linlearning} proposed an unsupervised deep learning approach to learn compact binary codes. Three criterions on binary codes are learned in their network: 1) minimal loss quantization, 2) evenly distributed codes, and 3) uncorrelated bits. Wu et al.~\cite{wu2017unsupervised} present an end-to-end unsupervised deep video hashing (UDVH) which integrates the feature clustering and feature binarization to preserve the neighborhood structure of the binary space, and a smart rotation to balance the binary codes. Xia et al.~\cite{xia2016unsupervised} proposed a novel unsupervised heterogeneous deep hashing framework, in which the auto-encoder and Restricted Boltzmann Machine (RBM) are utilized to learn the binary codes. Venkateswara et al.~\cite{venkateswara2017deep} introduced a new dataset called office-home to evaluate domain adaptation algorithms.

While, existing unsupervised hashing methods do not consider to leverage useful knowledge from the related datasets. Yang et al.~\cite{yang2016zero} proposed zero-shot hashing for encoding the images of the unseen classes to the binary codes. In their work, they firstly use NLP model to transform data labels into semantic embedding space, in which the relationships among the seen and unseen classes can be well characterized. Then, the embedding space is rotated for better aligning the visual feature space. Finally, hash functions are learned to transform visual feature space into the embedding space.

Since the underlying data distributions of the seen classes and the novel classes are different, the projection functions learned by the seen classes without any adaptation to the novel classes may cause data bias problem. Fu et al.~\cite{fu2015transductive} proposed transductive multi-view embedding to solve the projection domain shift problem. Guo et al.~\cite{socher2013zero} proposed transductive zero-shot recognition via jointly learning the shared model space for transferring the knowledge between the classes. The most similar work is \cite{pachori2017hashing}, which formulated it as the domain adaptation problem for zero-shot hashing. Given the features of a mini-batch of images belong to the unseen classes, it updates the transformation matrix learned from the seen classes in each iteration.

Although success, most existing zero-shot hashing methods firstly represent each input image by a fixed visual descriptors (e.g., extracted from the pre-train deep models~\cite{yang2016zero}), then followed by separate projection and quantization steps to encode the representation into a binary code. Such fixed visual features may not be optimal and it is expected to learn the visual features that can sufficiently preserve the images similarities. In this paper, we propose a deep framework for learning better common representation. 

\section{Transductive Zero-Shot Hashing}
\label{our_method}

In this section, we describe an architecture of deep convolution network designed for transductive zero-shot hashing (TZSH). 

We firstly introduce notations to formalise the TZSH setting. There is a labeled source dataset $S = \{I_i^s, Y_i^s\}_{i=1,\cdots,n}$, where $I_i^s$ is an image and $Y_i^s$ is the class name/label of the $i$-th image. We also have a large set of unlabeled data $U = \{I^u_i\}_{i=1,\cdots,n_u}$, which includes the images from both the seen classes $Y^s$ and the novel classes $Y^u$ but no annotations, where the novel classes $Y^u$ are disjoint from the seen ones, that is $Y^u \cap Y^s = \varnothing$. The goal of TZSH is to learn a deep model from the labeled images from the source dataset $S$ and the unlabeled dataset $U$, while the similarities among the novel images and seen images should also be preserved. 

As shown in Fig.~\ref{framework}, the proposed architecture mainly contains two parts: 1) a shared deep network for learning the common semantic image representations for both the source and target datasets, and 2) a coarse-to-fine similarity mining module for finding the  similarities of the target images with the help of the semantic word spaces. By the word representations of class names, we transfer the similarities from the seen classes to the novel ones by the proposed cross-images selection layer in a greedy manner. Finally, we can easily construct the loss function by using the above found similarities. 

\subsection{Common Semantic Representation via Shared Two-streams Network }
To encode both the source and target images into the same semantic space, we introduce a two-stream architecture, which includes two inputs. The first stream operates on the labelled source data $\{I_i^s, Y_i^s\}_{i=1,\cdots,r^s}$, and the second stream operates on the unlabelled data $\{I_i^u\}_{i=1,\cdots,r^u}$, where $r^s$ and $r^u$ are the sizes of mini-batch, respectively. These images go through the deep network with the stacked convolutional and full-connected layers, and are encoded to a common representation space. Note that the weights of the corresponding layers are shared between the two streams, and are trained jointly. The output common semantic representations are denoted as $\mathbf{f}^s_i = \psi(I_i^s)$ and $\mathbf{f}^u_i = \psi(I_i^u)$, where $\psi$ is the proposed deep network. 


\subsection{Coarse-to-Fine Similarity Mining}
\label{second_similar}
There are no annotations for the unlabelled images, making it a challenging problem for constructing the hash functions. With the learned common semantic spaces, we aim to detect the similarities among the unlabelled images by the proposed coarse-to-fine approach in this subsection. 

We use the following two stages to find the similarities among the target data: 1) we first greedily select the most informative images in the coarse stage, and then 2) we further greedily select an image from the coarse set for each of the target class, respectively. 

\begin{figure}[h]
\centering
    \includegraphics[width=1\hsize \hspace{0.01\hsize}]{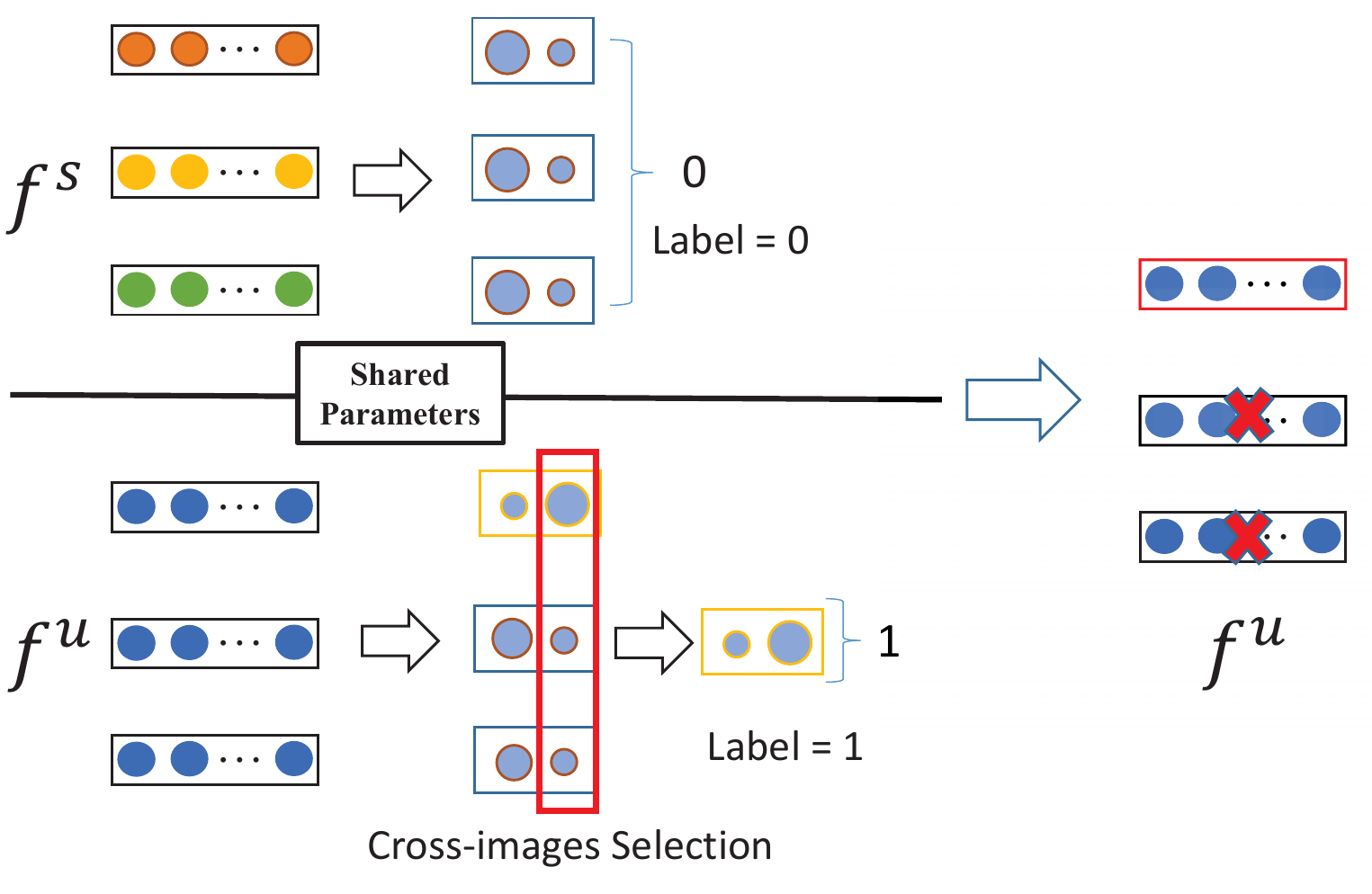}
  \caption{Coarse similarity mining module. To detect the most informative target images, we formulate it as a binary classification problem, in which images of the seen classes are regarded as the negative samples and images of the novel classes are positive samples. Without annotations of images, we propose a cross-images selection layer to find the positive samples. For all $r'$ images, the network generates a $2$-dimensional probability vector for each image, in which the first value represents the probability score coming from the seen classes and the second one represents the probability score from the novel classes. The cross-images selection layer greedily selects one image that has the highest possibility score, i.e., the maximum in the second values of all the $r'$ images. The selected image is regarded as the most informative image.
}
  \label{hash}  
\end{figure}

\subsubsection{Coarse Similarity Mining}




The main idea behind our method is that there is commonly a difference between the distributions of the images from the novel classes and the images from seen classes. Thus, we can formulate it as a binary classification problem: the images of the seen classes are the negative samples and that of the novel classes are the positive samples. To train the model, we can use the labeled images (all from the seen classes) as the negative samples. Due to no annotations, the greedy method is applied and the cross-image selection layer is proposed for finding the positive samples.

More specifically, we add a fully-connected layer with $2$-dimensional after the semantic representations $\mathbf{f}^u$ and $\mathbf{f}^s$, respectively. We have the deep features $\mathbf{c}^u \in \mathbb{R}^{r^u \times 2}$ and  $\mathbf{c}^s \in \mathbb{R}^{r^s \times 2}$, in which the $i$-th row of $\mathbf{c}^u$ represents the probability of the image coming from the seen classes is $\mathbf{c}^u_{i1}$ or from the novel classes is $\mathbf{c}^u_{i2}$. We propose a cross-images selection layer in our network for solving the target data without annotations. It greedily chooses the most representative images, e.g., $m$ images, which have the largest probability scores come from novel classes. Thus, we divide the $\mathbf{c}^u$ into $m$ groups with equal length $r'=r^u/m$, and each slice select one image which has the highest score. Fig.~\ref{hash} illustrates the proposed greedily selection procedure. Formally, we greedily select $m$ images that has the highest scores coming from the novel classes, which can be formulated as
\begin{equation}
\begin{split}
j_1 = arg \max \{ \mathbf{c}^u_{j2}, j=1,\cdots,r' \}, \  \cdots, \\
j_m = arg \max \{ \mathbf{c}^u_{j2}, j=(m-1)r'+1,\cdots,r^u \}.\\
\end{split}
\end{equation}
If the $j$-th image has a relatively higher/lower score, the $j$-th image likely/unlikely belongs to the novel classes. The highest score means most likely belong to the novel classes. Thus, these selected $m$ images are most informative images from the mini-batch images, we can use them as positive samples. 

To learn the parameters, the problem is a traditional binary classification problem. We can use the softmax loss~\cite{AlexNet} to optimize this problem, which is widely used loss function for classification problem in neural networks. The loss objective in coarse stage can be formulated as
\begin{equation}
 \min  - \frac{1}{m} \sum_{i=1}^{m} \log(\mathbf{c}^u_{j_12}) - \frac{1}{r^s}\sum_{i=1}^{r^s} \log(\mathbf{c}^s_{i1}).
 \label{unlabel_softmax}
\end{equation}

\textbf{Discussions.} First, we show that our proposed deep architecture is able to detect the novel images. Suppose that $j$-th image is not from the novel classes, it is of a seen class. In this case, the first stream will assign the low probability score to this image due to the images of the seen classes are regarded as negative samples and the network's parameters are shared. Thus, the probability score can be suppressed in the next iteration and make this image not to be chosen again. 

Second, the cross-images selection layer is attractive because it also allows end-to-end training. Specially, the gradient of the objective (\ref{unlabel_softmax}) is $\bf{g}$. Then, we can just put the gradient to the place where the image came from and let the gradients of other images to be zero, that is the gradients of informative images are $\bf{g}$ and other images are $\bf{0}$ in the cross-images selection layer when back propagation.




\subsubsection{Fine Similarity Mining}
\label{third_similar}
With the found $m$ images from the unlabelled data, we need a fine stage to find the similarities among these images. Although without the annotations, fortunately, the similarities among the class names are available, e.g., we can obtain them via the existing word2vec~\cite{mikolov2013distributed} model. Specially, suppose the $i$-th class name is $Y_i$, we denote $Z_i = g(Y_i)$ as the word-vector for the $i$-th class name. The similarity between the $i$-th class and the $j$-th class is defined as
\begin{equation}
Sim(Y_i,Y_j) = \frac{Z_i Z_j}{||Z_i|| \cdot ||Z_j||},
\label{sim}
\end{equation}
where $Sim(Y_i,Y_j)$ is the cosine similarity between the two vectors.

\begin{figure}
  \centering
  \includegraphics[width=1\hsize]{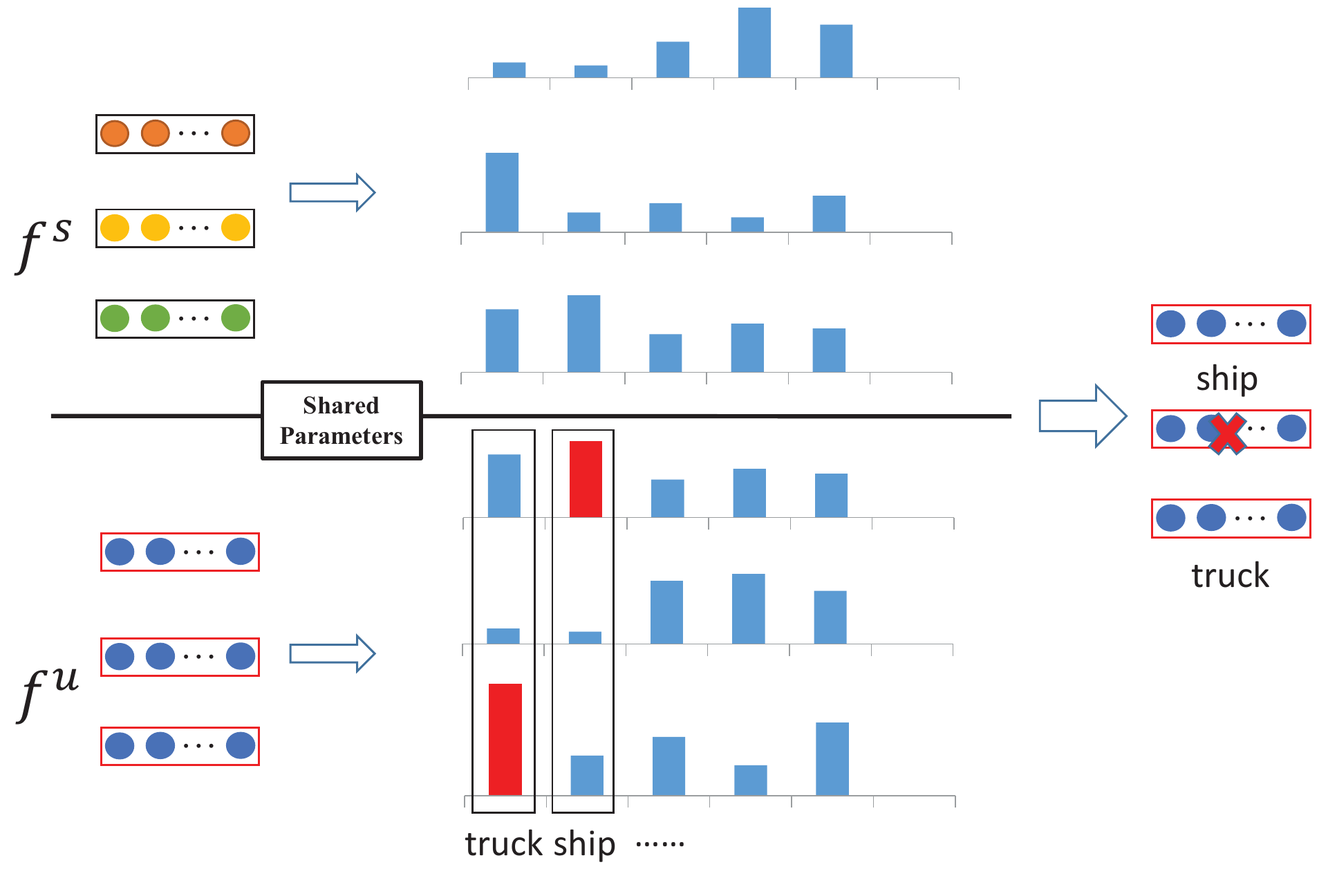}
  \caption{Illustration of the proposed fine similarity mining module. For the $m$ images (in $n_y$ novel classes), our network generates a probability vector for each images, e.g., $\mathbf{p}^u_i \in \mathbb{R}^{n_y}$ ($i=1,\cdots,m$). After that, the cross-images selection layer is used to select the $n_y$ images. Each image corresponds to one target class.} 
    \label{fine_similar}
\end{figure}

Inspired by that, we also use a greedy approach to detect the similarities among the target images. Fig.~\ref{fine_similar} shows the framework of our method. We project the source classes into the target classes. More specially, we add another fully-connected layer with $n_y$ dimensional features for $\{\mathbf{f}_i^s\}_{i=1,\cdots,r}$ and $\{\mathbf{f}_i^u\}_{i=j_1,\cdots,j_m}$, where $n_y$ is denoted as the number of target classes. These images go through the deep neural network and output the probability scores: $\mathbf{p}^u \in \mathbb{R}^{m \times n_y}$ and $\mathbf{p}^s \in \mathbb{R}^{r^s \times n_y}$. We also use the cross-images selection layer to greedy select the images which have the largest probability come from the novel classes:
\begin{displaymath}
\begin{split}
j_1 = arg \max \{ \mathbf{p}^u_{j1}, j=1,\cdots,m \}, \\
j_2 = arg \max \{ \mathbf{p}^u_{j2}, j=1,\cdots,m \}, \\
\cdots,\\
j_{n_y} = arg \max \{ \mathbf{p}^u_{jn_y}, j=1,\cdots,m \}.
\end{split}
\end{displaymath}

For training the parameters, we also use the softmax loss to optimize the problem. For the source data, we firstly calculate the similarities between the labels $Y^s_i$ and all target classes according to (\ref{sim}):
\begin{equation}
\hat{Sim}_i = [Sim(Y^s_i,Y_1^u),\cdots,Sim(Y^s_i,Y_{n_y}^u)].
\end{equation}
Then, a normalization process is calculated as to
make sure that $\sum_k \hat{Sim}_{ik} = 1$. $\hat{Sim}_i$ is the probability scores that the image looks like the novel classes, which is used as the label for the $i$-th image. For the target data, the selected images are assigned to the corresponding classes, respectively. The softmax loss can be defined as
\begin{equation}
\min - \frac{1}{r^s} \sum_{i=1}^{r^s} \sum_{j=1}^{n_y} \hat{Sim}_{ij} \log(\mathbf{p}^s_{ij}) - \frac{1}{n_y} \sum_{i=1}^{n_y} \log(\mathbf{p}^u_{j_ii})
\end{equation}


%

In the proposed fine similarity mining module, the performance depends on the similarities between the seen and novel classes. If the novel class has high similarity with one seen class, our method can work well. If the novel class is not related to any seen classes, our method may fail.

\subsection{Similarity-Preserving Loss}
As we aim to learn the $l$-dimensional binary codes, we use a fully-connected layer to compact each $\mathbf{f}$ into a $l$-dimensional vector $\mathbf{h}$. Through the coarse-to-fine module, we have the deep binary features $\{\mathbf{h}^s_i\}_{i=1,\cdots,r^s}$ and $\{\mathbf{h}^u_i\}_{i=j_1,\cdots,j_{n_y}}$ for the source data and the most representative target data. 

The hamming distance between the similar images should be small and the hamming distance between the dissimilar images should be large. For the source data, the similarities can be obtained via the classes $Y^s$, i.e, $\hat{s}_{ik} = 1$ if $Y^s_i = Y^s_k$, otherwise  $\hat{s}_{ik} = 0$. For the $i$-th image comes from the source classes and the $k$-th image from the target classes, the similarity is $\hat{s}_{ik} = 0$. For the similarities among the target images, we use both the hamming distance and detected similarities as the supervised information to make sure not introduce too much wrong annotations. Specially, if the indices of the largest value in $\mathbf{p}^u_{j_i}$ and $\mathbf{p}^u_{j_k}$ are the same and the hamming distance between $\mathbf{h}^{u}_{j_i}$ and $\mathbf{h}^{u}_{j_k}$ is small, the two images are similar, i.e., $\hat{s}_{ik} = 1$. When the indices are not the same and the hamming distance is large, they are dissimilar, i.e., $\hat{s}_{i k} = 0$. The overall loss function can be defined as 
\begin{equation}
\begin{aligned}
&\min \sum_{i=1}^{r^s + n_y} \sum_{k=1}^{r^s + n_y} \hat{s}_{ik} ||\mathbf{h}_i - \mathbf{h}_k||^2 & \\
&+ (1 - \hat{s}_{ik}) \max(0, \epsilon -||\mathbf{h}_i - \mathbf{h}_k||^2).& 
\end{aligned}
\label{siamese_loss}
\end{equation}

\section{Experiments}
\label{experiments}

In this section, we evaluate and compare the performance of the proposed method with several state-of-the-art algorithms.

\subsection{Datasets and Experimental Settings}
We conduct extensive evaluations of the proposed Transductive Zero-Shot Hashing (TZSH) on three benchmark datasets.
\begin{itemize}
\item \textbf{ImageNet~\footnote{http://image-net.org/challenges/LSVRC/2012/}} is a subset of the large hand-labeled dataset. It consists of about 1.2 million images and 1,000 categories. 



\item \textbf{Animals with Attributes (AwA)~\footnote{http://attributes.kyb.tuebingen.mpg.de/}} consists of 30,475 images of 50 animals classes, which 85 numeric attribute values are provided for each class. 

\item \textbf{CIFAR-10~\footnote{http://www.cs.toronto.edu/~kriz/cifar.html}} consists of 60,000 images in 10 classes, which each class has 6,000 images. It is labeled subset of the 80 million tiny images dataset. 

\end{itemize}
In each dataset, we resize images of all these databases into $256\times 256$.  We randomly select 2,500 images from the target classes as the test set, and the rest images are used as the retrieval database. In the retrieval database, 10,000 images from the seen classes are randomly chosen as the labeled set $S$ and the other images are used as unlabeled set $U$.

We evaluate the performance by the popular metric: Mean Average Precision (MAP)and the precision scores within Hamming distance 2 as the evaluation metric. 



We implement the proposed method using the open-source \textit{Caffe}~\cite{jia2013caffe} framework. The mini-batch sizes are setted to $r^s=128$, $r^u=256$, and $m$ is setted to 32 for all experiments. In this paper, we use AlexNet~\cite{AlexNet} as our basic network. For all deep learning-based methods, the weights of layers are initialized by the pre-trained AlexNet model~\footnote{http://dl.caffe.berkeleyvision.org/bvlc\_alexnet.caffemodel} unless noted otherwise. IMH~\cite{shen2013inductive}, COSDISH~\cite{kang2016column}, SDH~\cite{shen2015supervised}, ZSH~\cite{yang2016zero}, One-Stage Hashing~\cite{onestep} and Domain Adaptation Zero-shot Hashing (DA-ZSH)~\cite{pachori2017hashing} are selected as the baselines.


 

\subsection{Results on ImageNet}
ImageNet is a large dataset and consists of 1.2 million color images from 1,000 categories. For fair comparison, we following the setting of~\cite{yang2016zero}. Precisely, 100 classes that have corresponding semantic word vectors learned from Wikipedia text corpus are selected. We randomly select 10 categories as the target classes and the rest 90 categories as the seen classes. 

The pre-trained AlexNet model uses all 1,000 classes in training. It is not suitable for the setting of zero-shot hashing. Thus, we train a new AlexNet model only using the rest 900 classes. The weights of our network are initialized by the new model. And the new model is also used to extract features for the images. 

\begin{figure}[ht!]
  \begin{flushleft}
  \centering
  \subfigure[MAP]{\label{NUS-WIDE-a}
   \raisebox{-0.01cm}{
   \includegraphics[width=0.3\textwidth]{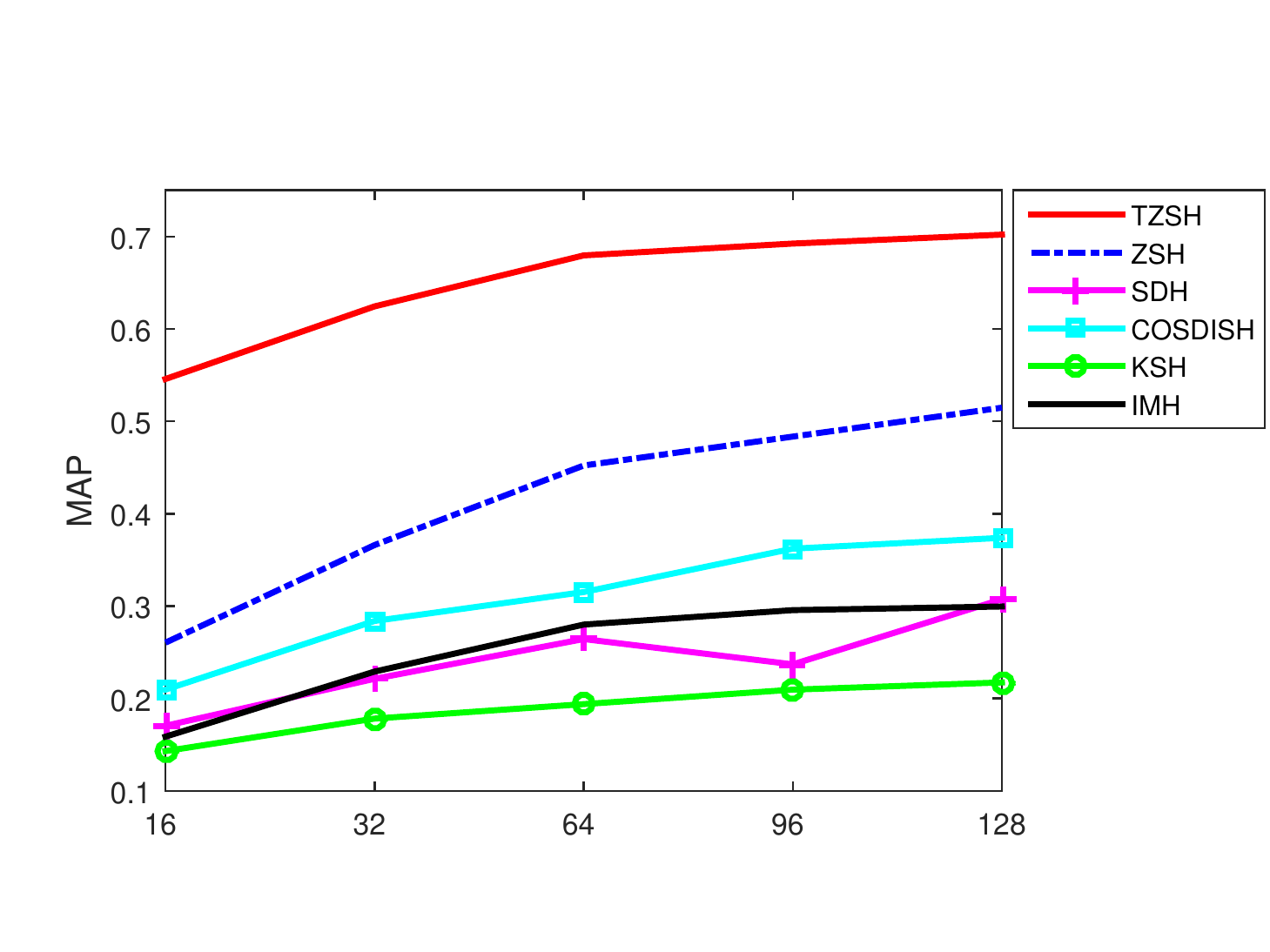}
  }}
  \subfigure[Precision scores within Hamming distance 2]{\label{NUS-WIDE-b}
  \raisebox{-0.01cm}{\includegraphics[width=0.3\textwidth]{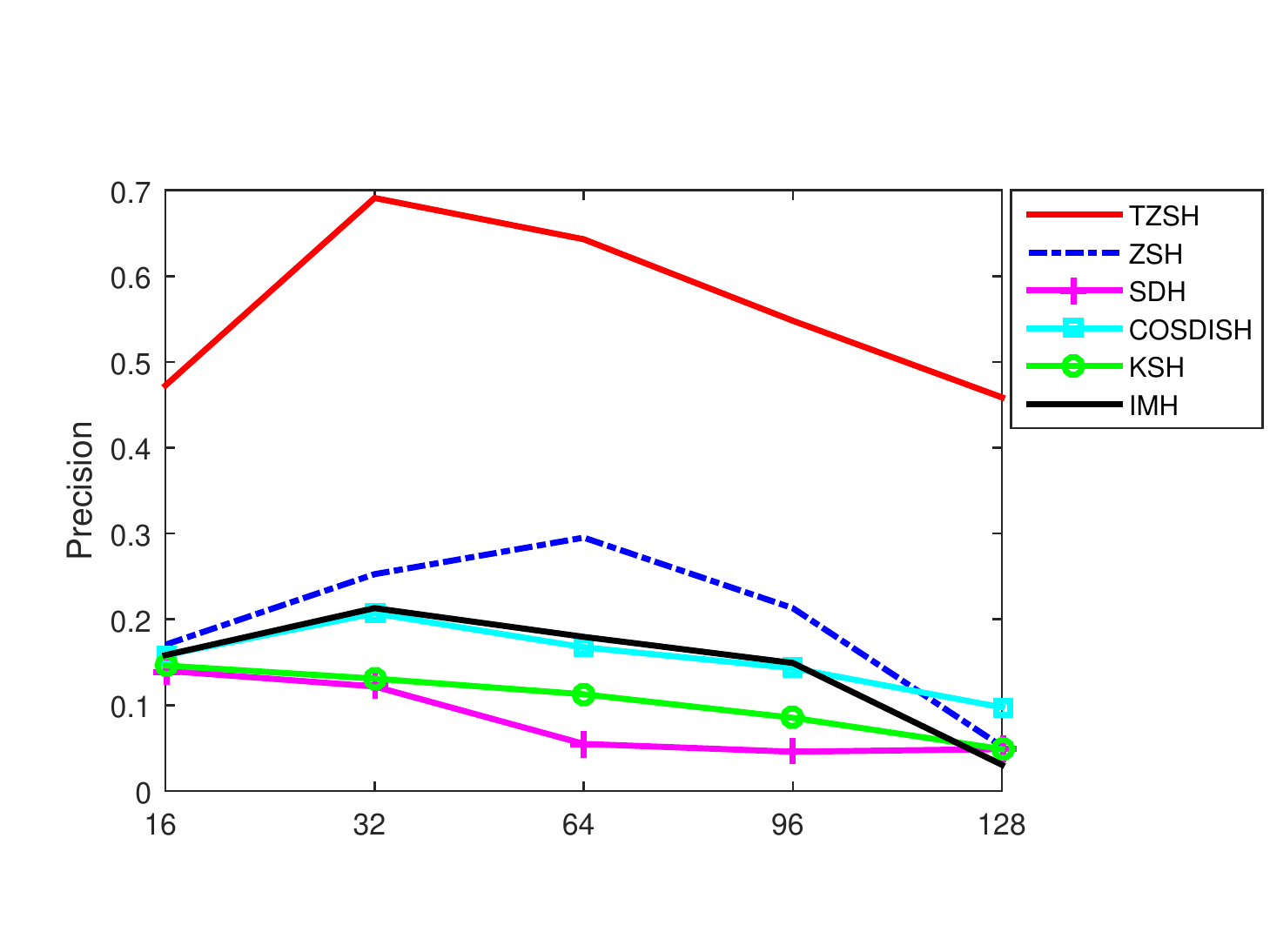}
  }}
  \caption{\footnotesize MAP and precision scores within Hamming distance 2 w.r.t. different numbers of bits on ImageNet dataset.}
  \label{imagenet-result}
  \end{flushleft}
\end{figure}

We compare the proposed method with several state-of-the art methods as shown in Fig~\ref{imagenet-result}. Note that the results of other methods are directly cited from the work~\cite{yang2016zero}. As can be seen, the proposed TZSH performs better than the baselines in ImageNet dataset. Note that ZSH is the recent proposed method, which achieves an excellent performance on this dataset. Even so, our method also performances better than ZSH.

\subsection{Results on AwA Dataset}

AwA dataset contains 50 animals species. For fair comparison, we following the setting of the most similar work domain adaptation zero-shot hashing (DA-ZSH)~\cite{pachori2017hashing}, which also uses the unlabeled data of the novel classes. Specially, 10 classes are selected as the target classes and the rest 40 classes as the seen classes. The 85 numeric attribute values for each class are used as semantic vectors. 

\begin{table}[h]
\small
    \centering \caption{MAP of Hamming ranking w.r.t different numbers of bits on AwA dataset.}
    \begin{tabular}{|c|c c c c c|}
        \hline
\multirow{2}{*}{ Method } & \multicolumn{5}{|c|}{AwA (MAP)} \\
& 16 bits & 32 bits & 64 bits & 96 bits & 128 bits  \\
        \hline
        IMH &  0.1082 & 0.1092 & 0.1102 & 0.1102 & 0.1103 \\
         \hline
         SDH & 0.0948 & 0.0743 & 0.1127 & 0.1266 & 0.1135  \\
         \hline
           COSDISH & 0.0779 & 0.1045 & 0.1184 & 0.1115 & 0.1035\\
         \hline
         	ZSH & 0.1087 & 0.1051 & 0.1219 & 0.1171 & 0.1208  \\
         \hline
           NDA-ZSH  & 0.0834 & 0.1067 & 0.1148 & 0.1278 & 0.1281\\
         \hline
           DA-ZSH & 0.1030 & 0.1261 & 0.1375 & 0.1432 & 0.1432 \\
         \hline
           TZSH & \bf{0.4010} & \bf{0.4803} &  \bf{0.5327} & \bf{0.5428} & \bf{0.5442}\\
          \hline
        \end{tabular}
    \label{map_awa}
\end{table}

The comparison results are shown in Table~\ref{map_awa}. Note that the results of other methods are directly cited from the paper~\cite{pachori2017hashing}, which also uses the features extracted from the deep network. Again, our method yields the highest accuracy and beats all the baselines. Note that we only train our model by the labelled images of the seen classes and the unlabelled data, which indicate that our model can performance well on the novel classes even without the annotations.

\subsection{Results on CIFAR-10}
This dataset consists of color images from 10 classes, i.e., airplane, automobile, bird, cat, deer, dog, frog, horse, ship and truck. For fair comparison, we follow the setting of the most similar work domain adaptation zero-shot hashing (DA-ZSH)~\cite{pachori2017hashing}. We enrich each class with 300-dimensional semantic word vectors with real number by the pre-trained word2vec model~\cite{mikolov2013distributed}. Two classes are randomly selected as target classes and the rest eight classes as seen classes. 

Table~\ref{map_cifar10_unseen} shows the performances of all comparing approaches. We take ship-truck, automobile-deer, dog-truck and cat-dog as target classes for zero-shot hashing, respectively. We can see that our method can achieve very high accuracy compared to the baselines. For instant, DA-ZSH achieves an excellent performance on this dataset. Even so, the MAP of TZSH is 0.5502 on average when the number of bits is 16, while the second best algorithm DA-ZSH is 0.4126.

The main reason for the good performance of our method is that TZSH can find the similarities of the target images via the proposed similarity mining module. Thus, similarities of target classes can be incorporated and help to obtain a good performance. However, for cat and dog, these two classes are dissimilar to the other eight classes, leading to hard to transfer the similarities of the seen classes to the novel ones. Thus it is not supervising that the results are poor.

\begin{table}[ht!]
\small
    \centering \caption{MAP w.r.t different numbers of bits on CIFAR-10 dataset.}
    \begin{tabular}{|c|c c c c c|}
        \hline
\multirow{2}{*}{ Method } & \multicolumn{5}{|c|}{CIFAR-10 (MAP)} \\
& 16 bits & 32 bits & 64 bits & 96 bits & 128 bits  \\
        \hline
          IMH & 0.1975 & 0.2035 & 0.2087 & 0.2172 & 0.2233 \\
         \hline
         SDH & 0.1715 & 0.2022 & 0.2136 & 0.2214 & 0.2396  \\
         \hline
           COSDISH & 0.1721 & 0.2036 & 0.2164 & 0.2270 & 0.2424\\
         \hline
         	ZSH & 0.1921 & 0.2168 & 0.2223 & 0.2430 & 0.2663   \\
         \hline
         	One-Stage & 0.2128 & 0.2462 & 0.2526 & 0.2632 & 0.2512\\
         \hline
           NDA-ZSH  &  0.2290 & 0.2018 & 0.1954 & 0.1927 & 0.1875\\
         \hline
           DA-ZSH & 0.4126 & 0.4313 & 0.4580 & 0.5325 & 0.5529 \\
         \hline
         \hline
          \bf{ship-truck} & 0.6644 & 0.6768 & 0.6848 & 0.6908 & 0.6828 \\
           
            \bf{dog-truck} & 0.6263 & 0.6370  & 0.6345 & 0.6395 & 0.6400\\
           
           \bf{ auto-deer} & 0.6212 & 0.6266 & 0.6359 & 0.6347 & 0.6405\\
           \bf{ cat-dog} &  0.3093  & 0.3221  & 0.3118  & 0.3102 & 0.3140 \\
         \hline
        \end{tabular}
    \label{map_cifar10_unseen}
\end{table}


To better understand our proposed method, we do further analysis in the following subsections. 

\subsubsection{Effect of Seen Category Ratio} In this set of experiments, we compare the performance of our method w.r.t different number of seen classes, in which the number of seen classes vary from 2 to 8~\footnote{Note that we do not give the results when the number of seen classes is one and nine. The reason is that we cannot construct hashing functions with only one seen class and we know all the annotations in transductive setting when the number of seen class is nine.}. 

\begin{figure}
  \centering
  \includegraphics[width=0.8\hsize]{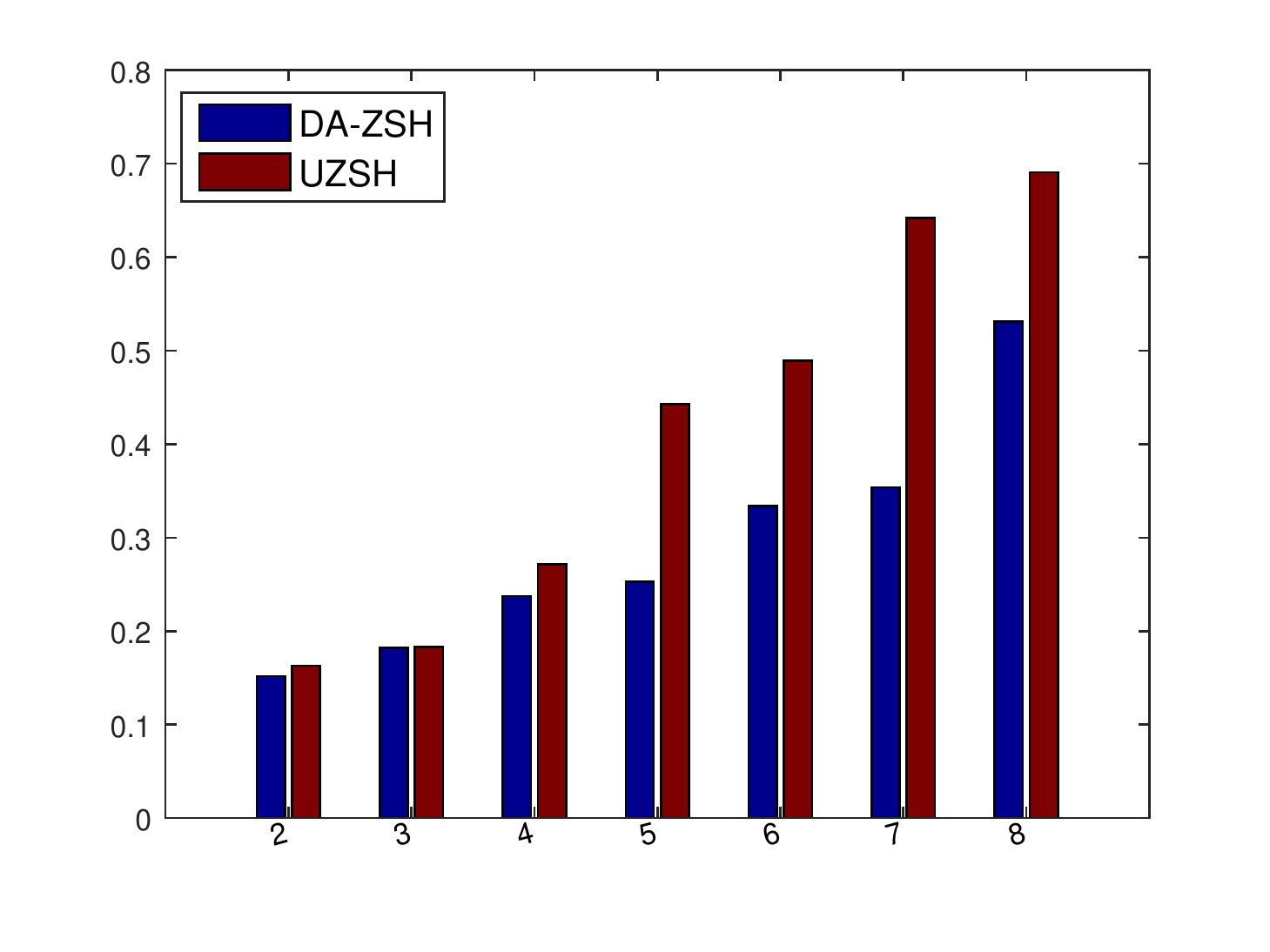}
  \caption{MAP on different number of the seen classes.} 
    \label{map_each_fig}
\end{figure}

Since the DA-ZSH~\cite{pachori2017hashing} is the most similar work and it performs very well, we choose it as the baseline. Fig~\ref{map_each_fig} shows the comparison results, from which it can be seen that: our method performs best in all different numbers of the seen classes. For example, the MAP of our method is 0.6908 compared to 0.5325 of the DA-ZSH when the number of the novel classes is two. We can also observe that our method achieves significant improvement over the baseline when the number of the seen classes is larger than the number of the novel ones. This is because it is hard to transfer the similarities of the seen classes to the novel ones when the number of the seen classes is small.

\subsubsection{Effect of the Unlabeled Size}
In training, we have an unlabeled set $U$. In this set of experiments, we explore the effects on MAP with respect to different numbers of the unlabeled size. We take ship-truck as target classes and the rest 8 classes as seen classes.

For exploring the effect of unlabeled size, the numbers of the unlabeled set are varying from 10,000 to 40,000. Table~\ref{unlabeled_size} shows the results. We can observe that our method can perform well on different numbers of the unlabeled set. This is because our method can detect the similarities among the ship-truck images.

\begin{table}[h]
\small
    \centering \caption{MAP of Hamming ranking w.r.t different numbers of unlabeled images on CIFAR-10 dataset.}
    \begin{tabular}{|c|c c c c c|}
        \hline
\multirow{2}{*}{ Method } & \multicolumn{5}{|c|}{MAP} \\
& 10,000 & 20,000 & 30,000 & 40,000 & 47,500  \\
         \hline
            TZSH & 0.6767 & 0.6727  & 0.6786 &  0.6833 & 0.6768\\
          \hline
        \end{tabular}
    \label{unlabeled_size}
\end{table}

\section{Conclusion}
\label{conclusion}

In this paper, we proposed a deep-network-based transductive zero-shot hashing method for image retrieval. In the proposed deep architecture, we designed coarse-to-fine similarity mining for finding the similarities of the novel classes. A deep binary classification network was proposed to find the most informative images from the target classes in the unlabelled data. Then, we transferred the similarities of the seen classes to the found images by the help of word representations. Based on the found similarities, we finally proposed a ranking loss function for preserving the similarities. Empirical evaluations on three datasets showed that the proposed method significantly outperforms the state-of-the-art methods.

In future work,we plan to study zero-shot hashing for multi-label image retrieval, in which the image may contain objects of multiple categories.

{\small
\bibliographystyle{ieee}
\bibliography{0812}
}

\end{document}